\documentclass[10pt,twocolumn,letterpaper]{article}

\usepackage{cvpr}
\usepackage{times}
\usepackage{epsfig}
\usepackage{graphicx}
\usepackage{amsmath}
\usepackage{amssymb}
\usepackage{multirow}
\usepackage{makecell}
\usepackage{algorithm}
\usepackage{algorithmic}
\usepackage{color}
\usepackage{caption}
\usepackage{booktabs}
\usepackage{subfigure}

\usepackage[breaklinks=true,bookmarks=false]{hyperref}

\cvprfinalcopy 

\def\cvprPaperID{****} 

\begin{document}

\title{Structure Learning of Deep Neural Networks  with Q-Learning}

\author{Guoqiang Zhong\\
{\tt\small gqzhong@ouc.edu.cn}
\and
Wencong Jiao\\
{\tt\small 939843754@qq.com}
\and
Wei Gao\\
{\tt\small gaoweichn@126.com}\\
Department of Computer Science and Technology, Ocean University of China\\
238 Songling Road, Qingdao, China 266100\\
}

\maketitle

\begin{abstract}
 Recently, with convolutional neural networks gaining significant achievements in many challenging machine learning fields, hand-crafted neural networks no longer satisfy our requirements as designing a network will cost a lot, and  automatically generating architectures has attracted increasingly more attention and focus. Some research on auto-generated networks has achieved promising results. However, they mainly aim at picking a series of single layers such as convolution or pooling layers one by one. There are many elegant and creative designs in the carefully hand-crafted neural networks, such as Inception-block in GoogLeNet, residual block in residual network and dense block in dense convolutional network. Based on reinforcement learning and taking advantages of the superiority of these networks, we propose a novel automatic process to design a multi-block neural network, whose architecture contains multiple types of blocks mentioned above, with the purpose to do structure learning of deep neural networks and explore the possibility whether different blocks can be composed together to form a well-behaved neural network. The optimal network is created by the Q-learning agent who is trained to sequentially pick different types of blocks. To verify the validity of our proposed method, we use the auto-generated multi-block neural network to conduct experiments on image benchmark datasets MNIST, SVHN and CIFAR-10 image classification task with restricted computational resources. The results demonstrate that our method is very effective, achieving comparable or better performance than hand-crafted networks and advanced auto-generated neural networks.
\end{abstract}

\section{Introduction}
During the last few years, deep learning has been playing an increasingly important role in the field of computer vision, such as image classification and object recognition, especially with the convolutional neural networks (CNNs) making great achievements. CNNs architectures evolving from traditional layer-stacking in a plain way like Alexnet \cite{alex}, VGG Net \cite{vgg}, to multi-branch Inception modules in GoogleNet \cite{inception}, shortcut connection in Residual Network (ResNet) \cite{resnet} and dense connection in Dense Convolutional Network (DenseNet) \cite{densenet}, have achieved increasingly high-performance. However, designing network architectures often not only needs a great many number of possible configurations, for example, the number of layers of each type and hyper-parameters for each type of layer, but also requires a lot of expert experience, knowledge and plenty of time. Hence there is a growing trend from hand-crafted architecture designing to automated network generating. Some research work \cite{nas,metaqnn,blockqnn,eas} has been done to automatically learn well-behaved neural network architectures and made promising results. Despite the learned networks have yielded nice results, the work in \cite{nas} and \cite{metaqnn} are just directly generate the entire plain network by stacking single layers one by one, while in \cite{blockqnn}, it aims at automatically generating block structure.

\begin{figure}[h]
  \centering
  \includegraphics[height=1.4in,width=3in,angle=0]{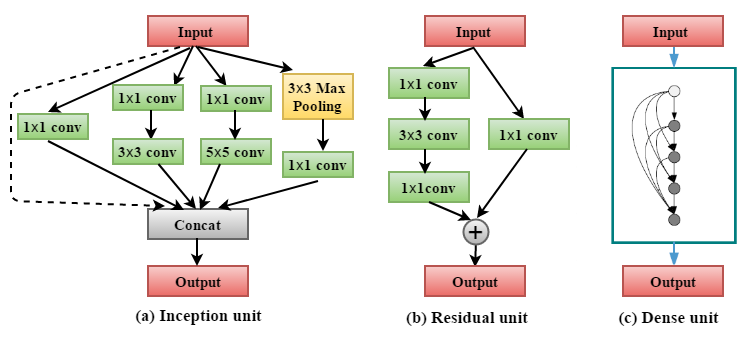}\\
  \caption{(a) Inception block unit of GoogLeNet (except for the dotted line). (b) Residual block unit of ResNet. (c) Dense block unit in DenseNet. The input shown in block units refers to the output from the last block, and the output would be the input of the next block. And more, a complete GoogLeNet, ResNet, DenseNet are stacked by their respective block units.}\label{figure1}
\end{figure}

\begin{figure}[h]
  \centering
  \includegraphics[height=1.4in,width=2.8in,angle=0]{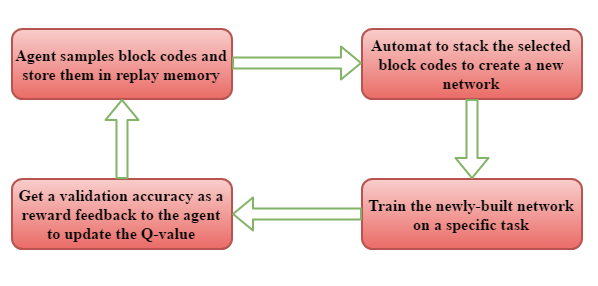}\\
  \caption{The overall model generation process: The Q-learning agent firstly samples block codes (the description of our blocks, introduced in detail later) and store them in replay memory, then according to a set of picked block codes to automatically stack blocks to construct a network and train it on specific tasks. Via training the newly-built network, we can receive a validation accuracy which can be regarded as the reward feedback to the Q-learning agent to update its Q-value. Consequently, the agent would be prompted to select a better-performing model in the next iteration.}\label{figure2}
\end{figure}

Starting from 2014, deeper and wider networks are utilized to significantly improve the performance of network architecture, emerging a number of networks that are no longer stacked layer by layer, among which the representative architectures include Inception network \cite{inception}, ResNet \cite{resnet} and DenseNet \cite{densenet}. These networks have many excellences that are deserved us to learn from and are crucial for learning good feature representations. For instance, the inception architecture is based on multi-scale processing; the residual unit in ResNet element-wisely adds the input features to the output by adopting skip connection to enable feature re-usage, while the dense block in DenseNet concatenates the input features with the output features to enable new feature exploration. Moreover, the architectures of these networks are mostly assembled as the stack of respective block units, which can been seen in Figure \ref{figure1}.
In this paper, unlike other methods \cite{nas,metaqnn,blockqnn,eas}, we are willing to make use of the benefits of these state-of-the-art blocks mentioned above and explore the possibility whether different blocks can be composed together to form a well-behaved neural network, so that we specify several different types of block modules and propose a novel automatic process to stack them to generate a multi-block neural network. We regard a block module as a layer of a CNN model to construct the whole network. And we learn the structures of deep networks based on these auto-generated and diverse multi-block neural networks. Moreover, we employ the well-known Q-learning \cite{qlearning} as an agent to sequentially select block modules of a CNN model with \(\epsilon\)-greedy strategy \cite{epsilon}. The validation accuracy on the given machine learning task would be viewed as the reward feedback to the agent to pick an architecture. By utilizing experience replay technique \cite{replay}, the goal of Q-learning agent is to find the optimal network architectures that behave well without manual intervention. Our model generation method can be summarized in Figure \ref{figure2}.

 We conduct experiments on standard image benchmark datasets: CIFAR-10, SVHN and MNIST for image classification tasks. The automatically constructed network architectures present competitive performance compared with those hand-crafted networks as well as auto-generated networks, achieving the best result on CIFAR-10, SVHN and MNIST with 4.68\%, 1.96\%, 0.34\% test error rate respectively. In addition to obtaining well-behaved multi-block neural networks via reinforcement learning (RL),  we have some interesting findings from our replay memory which stores a lot of data about structure information of multi-block networks.
\section{Related Work}

\subsection{Convolutional Neural Network Architecture Designing}
Traditional CNNs usually include convolution layers, pooling layers, and fully connected layers. These layers are then stacked to construct a deeper network. Starting from AlexNet \cite{alex} to the more modern DenseNet \cite{densenet}, neural network architectures have changed much to improve the performance of deep CNNs, and promising results have been made on many machine learning tasks including image classification. There are also many studies on automating neural network design, from early works based on the genetic algorithm or other evolutionary algorithms \cite{evolving,genetic}, and Bayesian optimization \cite{bayes}, whose performance, however, could not be comparable to the hand-crafted networks as far as we know, to recent works, like Neural Architecture Search (NAS) \cite{nas}, Meta-QNN \cite{metaqnn} and Block-QNN \cite{blockqnn}, using reinforcement learning method, achieving comparable or even higher performance.

The Meta-QNN presented in \cite{metaqnn} used Q-learning agent to sequentially pick CNN layers. While NAS proposed in \cite{nas} utilized an auto-regressive recurrent network as the agent to generate model descriptions which designate the architecture of a neural network and trained the recurrent network with policy gradient. Both of these two methods just directly create the entire plain network by stacking layers one by one based on reinforcement learning. While Block-QNN, introduced in \cite{blockqnn}, mainly aimed at automatically generating the block structure and then the optimal block was stacked repeatedly with several convolution and pooling layers to form the whole network with manual intervention.

Since most of the state-of-the-art neural networks are stack-based structures, we draw on this common pracitce to construct networks by stacking multiple types of blocks in an automatic way. We set up three types of block modules: dense block, residual block and inception-like block in our experiment. Each block type contains several sub-selections to increase the diversity of block choices, therefore the combination of blocks and the depth of the combined network are diverse.

\subsection{Reinforcement Learning}
In the problem setup of automatically generating an architecture of a CNN model, based on RL, we formulate the automatic architecture generating procedure as a sequential decision making process, where the state is the current network architecture and the action is to pick the successive block module; the agent interacts with the environment by performing a sequence of actions, namely, sequentially selecting block modules. After T steps selection, the final whole network is constructed by stacking theses block modules, and then trained in a specific task to get a validation accuracy, which would be returned as the reward to update the agent. In this paper, we use Q-learning method for updating the agent who is trained to maximize the cumulative reward. Along with \(\epsilon\)-greedy strategy and experience replay technique, the agent would be prompted to select a better-performing model in the next iteration.\\
\begin{figure*}[htbp]
  \subfigure[dense block]{
  \begin{minipage}{2cm}
  \centering
  \includegraphics[height=1.8in,width=0.6in,angle=0]{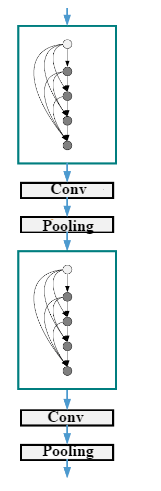}\\
  \end{minipage}}
  \subfigure[3 types of residual blocks]{
  \begin{minipage}{4.8cm}
  \centering
  \includegraphics[height=1.8in,width=1.9in,angle=0]{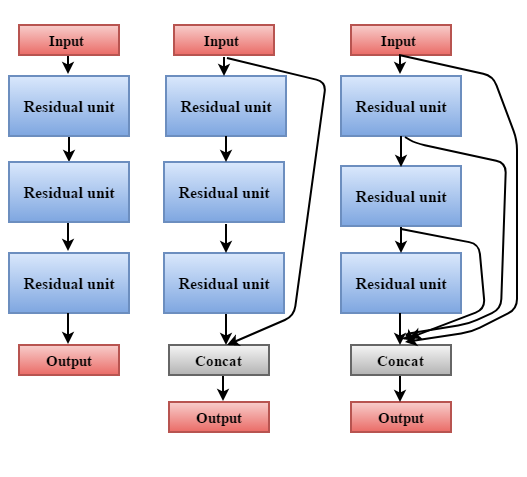}\\
  \end{minipage}}
  \subfigure[inception-like blocks]{
  \begin{minipage}{6.7cm}
  \centering
  \includegraphics[height=1.8in,width=4.2in,angle=0]{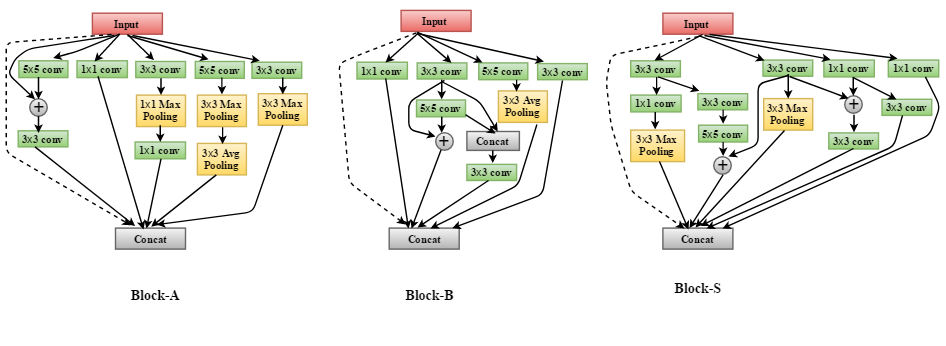}\\
  \end{minipage}}
  \caption{The block patterns used in our experiment. (a) shows our dense block that belongs to the first two blocks of the DenseNet (L = 40, k = 12) including their transition layers. (b) shows our different settings on residual blocks, which are based on residual unit in ResNet. (c) shows the inception-like blocks which add the concatenation on the original Block-A, Block-B, and Block-S to be as our block moules ( with the dotted line), and the block in Figure \ref{figure1} (b) with or without a dotted connection are also one of our inception-like blocks. }\label{figure3}
\end{figure*}

\section{Methodology}

\subsection{Block representation}
Considering the superior feature learning ability of DenseNet, we impose restrictions on the agent that it can only pick dense block in the first step, namely, the dense block is set as the start point of a network, and from the second action, it can select any other block modules according to the strategy until the whole network arrives at the terminate state. Note that, motivated by \cite{identity}, the convolution operation used in each block module refers to a composite function of three consecutive operations: traditional convolution (Conv), batch-normalization (BN) \cite{bn} and ReLU \cite{relu}, except that in dense block it is BN-ReLU-Conv according to \cite{densenet}.

\begin{figure}[h]
  \centering
  \includegraphics[height=1.6in,width=3.2in,angle=0]{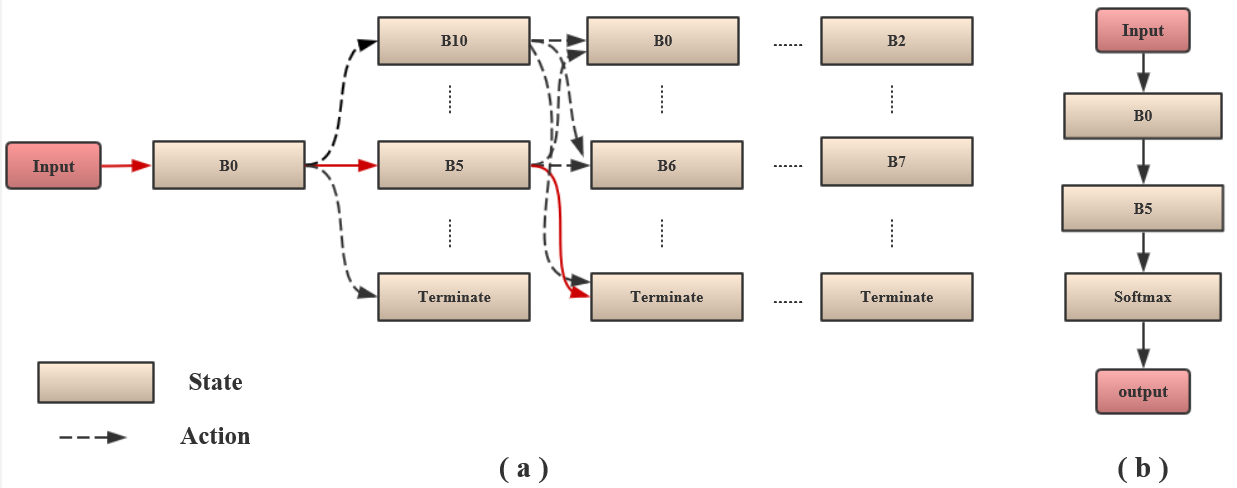}\\
  \caption{The state transition process by different action selections. In this illustration, the action selections in dotted line follow the \(\epsilon\)-greedy strategy in experiments. The red solid line in (a) shows a path that the Q-learning agent may choose along with the network structure defined in (b). Zoom in for a better view.}\label{figure4}
\end{figure}

One important point that needs to be explained is the blocks generated in advance by us. On account of that down-sampling is an essential part that can change the size of feature maps, whereas there is no pooling operation in our setup, we select the first two blocks of the DenseNet (L = 40, k = 12; L, k respectively refers to the depth and growth rate of DenseNet) \cite{densenet} including their transition layers as our dense block module (Figure \ref{figure3} (a)). By adopting a part of the well-structured DenseNet to ensure computational efficiency is another reason we considered. As to our residual blocks (Figure \ref{figure3} (b)) set according to the paradigm of the residual unit (Figure \ref{figure1} (a)) in ResNet, they contain three residual units. In addition to setting different numbers of filters, a slight change made to the original residual form is that we concatenate the input features of a block with the output of each residual unit or only with the output of the final residual unit, thus to impel our residual blocks not only reuse features but also explore new ones. When it comes to the inception-like blocks (Figure \ref{figure1} (b), Figure \ref{figure3} (c)), the setup are the same as the residual blocks. We also employ the Block-A, B, S proposed in \cite{blockqnn} due to their impressive performance, and make some changes on them by concatenating the input features with their output (Figure \ref{figure3} (c) with the dotted line).
\begin{table*}[htb]
\caption{Block Code State Search Space. This space includes 12 types of  block modules used in our approach. A selected neural network architecture can be seen in Figure\ref{figure4} (b).}\label{table1}
\centering
  \begin{tabular}{|c|c|c|}
  \hline
  Block Module& Choice\\
  \hline
  Dense Block& \makecell[l] {Only one dense block \textbf{B(0)} (shown in Figure \ref{figure3} (a)) can be selected.} \\
  \hline
  Residual Block& \makecell[l] {There are four selections: \\\textbf{B(1)} (shown in Figure \ref{figure3} (b), left) keeps the original residual form; \\\textbf{B(2)} and \textbf{B(3)} (shown in Figure \ref{figure3} (b), middle) are the variants that concatenate the input \\features only with the output of the final residual unit; \\\textbf{B(4)} (shown in Figure \ref{figure3} (b), right) concatenates the input features with the output of each \\residual unit.} \\
  \hline
  Inception-like Block& \makecell[l]{ There are seven selections: \\\textbf{B(5), B(6), B(7)} (shown in Figure \ref{figure1} (b) without the dotted line) are the original inception \\module form with different output channels;\\ \textbf{B(8)} (shown in Figure \ref{figure1} (b) with the dotted line) concatenates the input of the original \\inception module with its output;\\\textbf{B(9), B(10), B(11)} (shown in Figure \ref{figure3} (c) with the dotted line) are the variants of Block-A,\\ Block-B, and Block-S.} \\
  \hline
  Terminate&\makecell[l]{Global Avg. Pooling (\textbf{GAP)} / Softmax (\textbf{SM)} }\\
  \hline
  \end{tabular}
\end{table*}
In order to make it easier to represent the state space of our problem, we convert the structural information into block code, which can be seen as a simple 1-D vector for different blocks, as shown in Table \ref{table1}. And the state transition process where there exists a sequence of action choices can be depicted in Figure \ref{figure4}.

\subsection{Multi-Block Neural Network Design with Q-Learning}
Based on RL, whose goal is to maximize the cumulative reward according to how an agent takes actions, we employ Q-learning method \cite{qlearning} as the agent to sequentially select blocks of a CNN model with \(\epsilon\)-greedy strategy. The automatically generating architecture procedure can be seen as a sequential decision making process, and we model the block picking process as a Markov decision process (MDP), with the assumption that a well-performing block in one selected network should also perform well in another selected network.

The Q-learning model involves an agent, a state space, an action space and a reward function. As mentioned earlier, seeing a block as a layer of a CNN model, we define each state as a simple 1-D vector block code to describe the relevant block information. The state \(s\in S\) represents the status of the current layer, and the action \(a\in A\) is the decision for the next successive layer. For each state, there is a finite actions \(A(s)\subseteq A\)  that the agent can select from. Due to the defined block code, we can constrain the environment to be finite, discrete and relatively small so that the agent could deterministically terminate in a finite number of time steps. The state transition process \((s_t,a(s_t))\rightarrow(s_{t+1})\) is shown in Figure \ref{figure4} (a). The given task of the agent is to sequentially select the block code of a network, which can be seen as an action selection trajectory \emph{T}$_\emph{i}$, whose total expected reward is:\\
\begin{equation}\label{1}
 R_{T_i} = \sum_{(s,a,s¡¯ )\in {T_i}} E_{r|s,a,s¡¯}[ r|s,a,s¡¯]
\end{equation}

The agent's goal is to maximize the cumulative reward over all possible trajectories, \(max_{T_i\in T} R_{T_i}\). To solve this maximization problem, we utilize recursive Bellman Equation as follows:\\
\begin{equation}\label{2}
\begin{aligned}
   Q^*(s_t,a) &=  E_{ s_{t+1}|s_t,a} [E_{r|s_t,a,s_{t+1}}[ r|s_t,a,s_{t+1}] \\
   &+ \gamma max_{a¡¯ \in A(s_{t+1})} Q^*(s_{t+1},a¡¯) ]
\end{aligned}
\end{equation}
where at time step \(t\), given the current state \(s_t\), the agent takes a subsequent action \(a_t\in A(s_t)\), arrives in next state \(s_{t+1}\) and receives reward \(r_{t+1}\), then the maximum total expected reward is the state-action value \(Q^*(s_t,a)\), known as Q-values.
For the above quantity, it can be solved by formulating as an iterative update:\\
\begin{equation}\label{3}
\begin{aligned}
   Q_{new} (s_t,a_t) & \leftarrow (1- \alpha)Q_{old}( (s_t,a_t)\\
   &+ \alpha [ r_{t+1}+ \gamma max_a Q (s_{t+1},a) ]
\end{aligned}
\end{equation}
where Q-learning rate \(\alpha\)  determines the weight given to the the newly acquired information compared with the old information to ensure the stability of the learning process and the convergence in the final stage, discount factor \(\gamma\) determines the importance of the future rewards.

We employ Q-learning method to train the agent with experience replay technique to store the past explored multi-block network information and validation accuracy for fast convergence, and with \(\epsilon\)-greedy strategy to take random actions with the probability \(\epsilon\) and take greedy action with the probability 1-\(\epsilon\) for converging rapidly as well. By gradually reducing \(\epsilon\) from 1 to 0.1, the agent can start with an exploration stage, then transform to the exploitation stage such as to find more accurate networks. Table \ref{table2} shows the \(\epsilon\) schedule we set.
\begin{table}[h]
\caption{\(\epsilon\) schedule: at different \(\epsilon\) stages, the Q-learning agent trains different numbers of unique models according to the setting.}\label{table2}
\scalebox{0.75}[0.75]{%
\begin{tabular}{|c|c|c|c|c|c|c|c|c|c|c|}
  \hline
  \(\epsilon\)& 1.0&0.9&0.8&0.7&0.6&0.5&0.4&0.3&0.2&0.1 \\
  \hline
  \makecell[l]{Trained model\\numbers }&50&7&7&7&10&15&15&15&15& 20 \\
  \hline
\end{tabular}}
\end{table}

\section{Experiments}
\subsection{Q-learning Training Details}
During the Q-learning update iteration, which can be seen in Equation \ref{3}, we set the Q-learning rate \(\alpha\)  as 0.01 and discount factor \(\gamma\) as 1 to value the long-term rewards as much as the short-term rewards. As mentioned before, we utilized the replay memory called replay-database to store the network structure information and the accuracy on the validation set of all sampled models. And we used Adam optimizer \cite{Adam} with \(\beta\)$_{1}$ = 0.9, \(\beta\)$_{2}$ = 0.999, \(\epsilon\) = $10^{-8}$, to train each model for 30 epochs in total. The initial learning rate was set to 0.001, and it would be reduced by a factor of 0.2 every 5 epochs if the selected model started with a better performance than a random prediction, or otherwise it would be reduced by a factor of 0.4 and the model would be retrained with a maximum of 5 times. We utilized the MSRA initialization \cite{msra} to initialize all weights. The batch size on the training and validation sets was set to 64 and 50 respectively, and the maximum layer index for a multi-block network was set to 5 due to our restricted computational resource.

After the entire Q-learning process was completed, we selected the top 10 models on each dataset to do further training. Our experiments were implemented under the Caffe \cite{caffe} scientific computing platform, took 10-12 days to complete each dataset using a NVIDIA TITAN X GPU and a NVIDIA 1080 Ti GPU.

\subsection{Image datasets}
\textbf{CIFAR-10.} This dataset contains 50000 training images and 10000 test images which are colored with $32\times32$ pixels in 10 classes. We hold out 5000 random training samples as a validation set. And we preprocessed each image with global contrast normalization. In further training phase, we used Nesterov optimizer \cite{nesterov} with momentum, weight decay rate and initial learning rate set as 0.9, 0.0005 and 0.1 respectively to train the model 200 epochs in total, decreasing the learning rate to 0.01 at 100-th epoch and 0.001 at 150-th epoch. The batch size setting and the weight initialization were the same as that in Q-learning training process. \\
\textbf{SVHN.} The street view house numbers (SVHN) dataset consists of $32\times32$ colored digit images in 10 classes with 73257 training images, 26032 testing images and 531131 additional images for extended training. We only used the original training set and randomly picked 5000 training samples for validation, by preprocessing with local contrast normalization according to \cite{maxout}. After only training 30 epochs during the Q-learning process, there are nearly a tenth of all the selected models easily achieving an accuracy higher than 95\%. When doing further training, we follow the common practice to use all the original and extended training images, holding out 400 samples per class from original training set and 200 samples per class from the extended training set to make up our validation set. Then we use Nesterov optimizer \cite{nesterov} with momentum, weight decay rate and initial learning rate set as 0.9, 0.0005 and 0.01 respectively to train the model 40 epochs in total, decreasing the learning rate to 0.001 at 20-th epoch and 0.0001 at 30-th epoch to do further training. The batch size setting and the weight initialization weren't changed. \\
\textbf{MNIST.} This handwritten digit dataset has 10 classes with 60000 training images and 10000 testing images which are $28\times28$ pixels. We made global mean subtraction for prepossessing and took 10000 training images as a validation set. Without imposing any further training, the number of models whose prediction accuracy are over 99\% accounts for nearly one-seventh of the total. So we didn't push further training on this dataset.\\
\begin{table*}[htb]
\caption{Error rate(\%) comparison with state-of-the-art hand-crafted and auto-generated CNNs on MNIST, SVHN and CIFAR-10 datasets. The CIFAR-10$^{+}$ indicates data augmentation(translation and /or mirroring), while the results on SVHN and MNIST without any data augmentation. The results show that, our method can achieve comparable or even better performance compared with the listed CNNs no matter on which image dataset.  }\label{table3}
\centering
\scalebox{0.85}[0.85]{%
\begin{tabular}{c|c|c|c|c}
  \hline
  Method&MNIST & SVHN& CIFAR-10 & CIFAR-10$^{+} $\\
  \hline
  Network in Network\cite{nin}  & 0.47& 2.35 & 10.41 & 8.81 \\
  Highway Network\cite{highway}& - & -  & - & 7.72 \\
  FitNet\cite{fitnet}  & 0.51 & 2.42 & - & 8.39\\
  VGG\cite{vgg} & -  & - & - & 7.25\\
  ResNet(L=20)\cite{resnet} & - & - & -  & 8.75\\
  ResNet(L=32)\cite{resnet}& -& - & - & 7.51  \\
  ResNet(L=44)\cite{resnet} & -  & - & -& 7.17 \\
  ResNet(L=56)\cite{resnet} & -  & - & -& 6.97 \\
  ResNet(L=110)\cite{resnet1}  & -& 2.01& 13.63 & 6.41 \\
  DenseNet(k=12,L=40)\cite{densenet}  & -& \textbf{1.79} & 7.00 & 5.24\\
  \hline
  \hline
  MetaQNN(ensemble)\cite{metaqnn} & \textbf{0.32} & 2.06 & - & 7.32\\
  MetaQNN(top model)\cite{metaqnn}   & 0.44& 2.28& - & 6.92 \\
  NAS v1\cite{nas} & -  & - & -& 5.50 \\
  NAS v2\cite{nas} & - & - & - & 6.01 \\
  EAS\cite{eas}  &- & 1.83&- & 4.89 \\
  Multi-block(Ours) &  0.34 & 1.96 & \textbf{ 6.2}& \textbf{4.68} \\
  \hline
\end{tabular}}
\end{table*}
\subsection{Results}
Our best results on CIFAR-10 is 4.68\% error rate, namely, 95.32\% accuracy, achieving competitive performance with state-of-the-art hand-crafted and auto-generated networks. Similarly,  when it comes to the result on SVHN and MNIST, the lowest test error rates are 1.96\% and 0.34\% respectively, on bar with or even better than the best results of the previous models. The comparison on three datasets is shown in Table \ref{table3}. And the Table \ref{table4},  \ref{table5}, \ref{table6} list the top 10 model architectures selected by the Q-learning agent along with their prediction accuracy and total numbers of parameters on CIFAR-10, SVHN and MNIST respectively. From these tables we can see that,  the accuracy of our top 10 models are above 92.50\% on CIFAR-10; the test error rates of our listed top 10 models on SVHN are below 2.65\%; there are at least 10 models that are more accurate than 99.50\% on MNIST.\\
\begin{table}[h]
\caption{The top 10 model architectures selected on CIFAR-10 dataset by the Q-learning agent along with their prediction accuracy and total number of parameters, and the order refers to the iteration at which  the model is chosen by the agent.}\label{table4}
\scalebox{0.8}[0.8]{%
\begin{tabular}{|l|c|c|c|}
  \hline
  Net& Accuracy(\%) & Order & Parameters \\
  \hline
  [B(0),B(0),SM(10)] & 95.32 & 131 & 4.32M \\
  \hline
  [B(0),B(0),B(10),B(0),SM(10)] &93.34&95&7.37M \\
  \hline
  [B(0),B(6),B(7),SM(10)] & 93.28 & 156 & 5.27M \\
  \hline
  [B(0),B(0),B(2),B(2),SM(10)] &93.16& 124 & 22.17M \\
  \hline
  [B(0),B(0),GAP(10),SM(10)]& 92.94 & 133 & 4.29M \\
  \hline
  [B(0),B(10),SM(10)] & 92.92 & 15 & 5.41M \\
  \hline
  [B(0),B(3),B(4),B(0),SM(10)] & 92.86 & 115 & 7.54M \\
  \hline
  [B(0),B(0),B(6),B(0),SM(10)] & 92.72 & 151 & 7.67M \\
  \hline
  [B(0),B(0),B(3),B(9),SM(10)] & 92.58& 101 &2.74M \\
  \hline
  [B(0),B(8),B(3),B(0),SM(10)] & 92.50 & 118& 7.17M \\
  \hline
\end{tabular}}
\end{table}
\begin{table}[h]
\caption{The top 10 model architectures selected on SVHN dataset by the Q-learning agent along with their prediction accuracy and total number of parameters,  and the order refers to the iteration at which  the model is chosen by the agent.}\label{table5}
\scalebox{0.78}[0.78]{%
\begin{tabular}{|l|c|c|c|}
  \hline
  Net& Accuracy(\%) & Order & Parameters \\
  \hline
   [B(0),B(0),B(4),B(2),SM(10)] & 98.04 &160 &14.06M\\
  \hline
  [B(0),B(4),B(3),B(0),SM(10)]& 97.64&6 &7.67M \\
  \hline
   [B(0),B(0),SM(10)]&97.56& 19 & 4.32M \\
  \hline
   [B(0),B(9),B(0),B(4),SM(10)]& 97.44& 60 &5.47M \\
  \hline
  [B(0),B(3),B(2),GAP(10),SM(10)] &97.42&76&13.29M \\
  \hline
  [B(0),B(11),B(0),B(2),SM(10)] & 97.38& 97 &14.74M \\
  \hline
  [B(0),B(3),GAP(10),SM(10)] & 97.36& 37 & 4.39M\\
  \hline
  [B(0),B(3),B(5),B(0),SM(10)] & 97.36& 109&7.59M \\
  \hline
  [B(0),B(9),B(0),SM(10)]& 97.34&124 & 4.78M\\
  \hline
  [B(0),B(3),B(0),SM(10)] & 97.32& 65& 6.71M \\
  \hline
\end{tabular}}
\end{table}
\begin{table}[h]
\caption{The top 10 model architectures selected on MNIST dataset by the Q-learning agent along with their prediction accuracy and total number of parameters,  and the order refers to the iteration at which  the model is chosen by the agent.}\label{table6}
\scalebox{0.8}[0.8]{%
\begin{tabular}{|l|c|c|c|}
  \hline
  Net& Accuracy(\%) & Order & Parameters \\
  \hline
  [B(0),GAP(10),SM(10)] & 99.66&5 & 2.07M\\
  \hline
  [B(0),B(3),B(4),B(0),SM(10)]& 99.64&146 & 7.54M \\
  \hline
  [B(0),B(0),SM(10)]&99.60& 35 & 4.29M \\
  \hline
 [B(0),B(3),B(0),B(0),SM(10)] &99.58& 61 & 8.89M \\
  \hline
  [B(0),B(0),B(4),B(10),SM(10)] & 99.56&69& 4.32M \\
  \hline
  [B(0),B(2),B(8),B(0),SM(10)]& 99.56 &147 &13.55 M \\
  \hline
 [B(0),B(0),GAP(10),SM(10)]& 99.54&84 & 4.29M\\
  \hline
 [B(0),B(5),B(8),B(0),SM(10)] & 99.5 & 128& 5.48M \\
  \hline
  [B(0),B(3),GAP(10),SM(10)] & 99.5 & 18& 4.39M \\
  \hline
  [B(0),B(9),B(11),B(0),SM(10)] & 99.5 &152 &6.74M\\
  \hline
\end{tabular}}
\end{table}

\begin{figure}[h]
  \centering
  \subfigure[]{
  \begin{minipage}{4cm}
  \centering
  \includegraphics[height=1.5in,width=1.7in,angle=0]{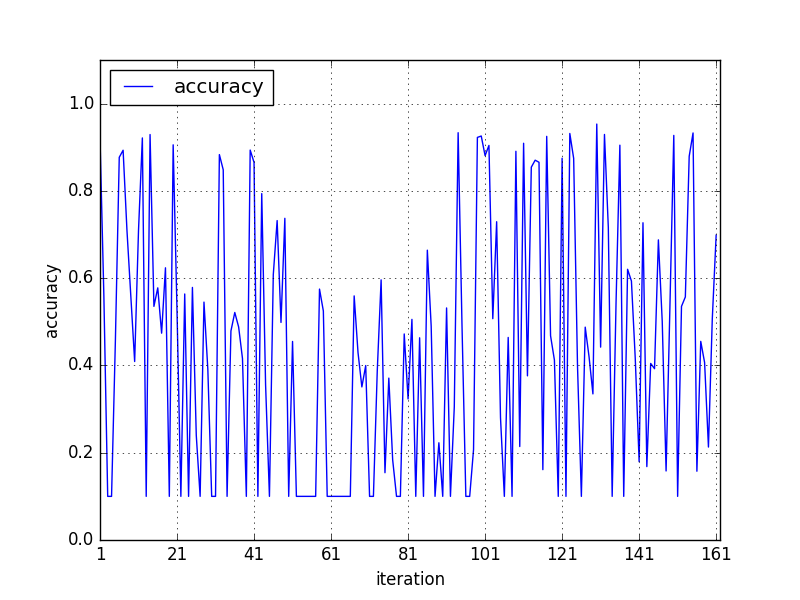}\\
  \end{minipage}}
  \subfigure[]{
  \begin{minipage}{4cm}
  \centering
  \includegraphics[height=1.5in,width=1.7in,angle=0]{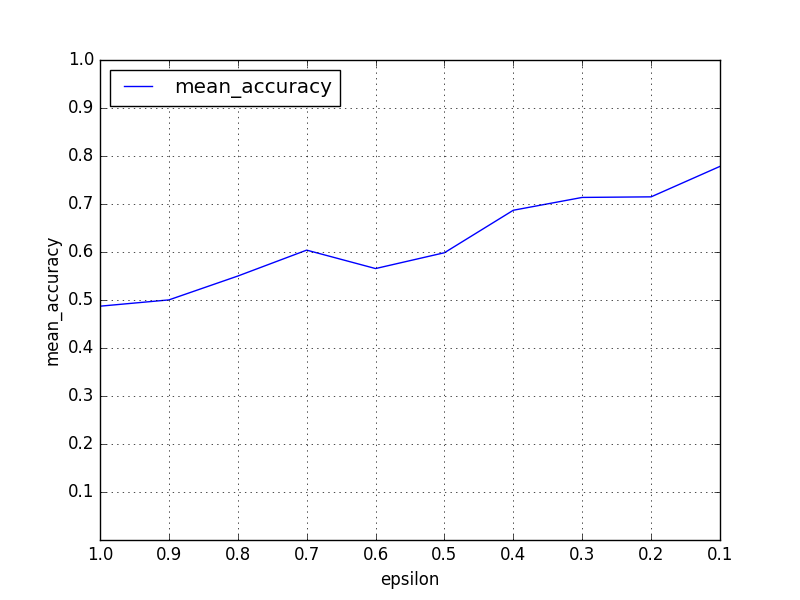}\\
  \end{minipage}}
  \quad
  \subfigure[]{
  \begin{minipage}{3.8cm}
  \centering
  \includegraphics[height=1.5in,width=1.7in,angle=0]{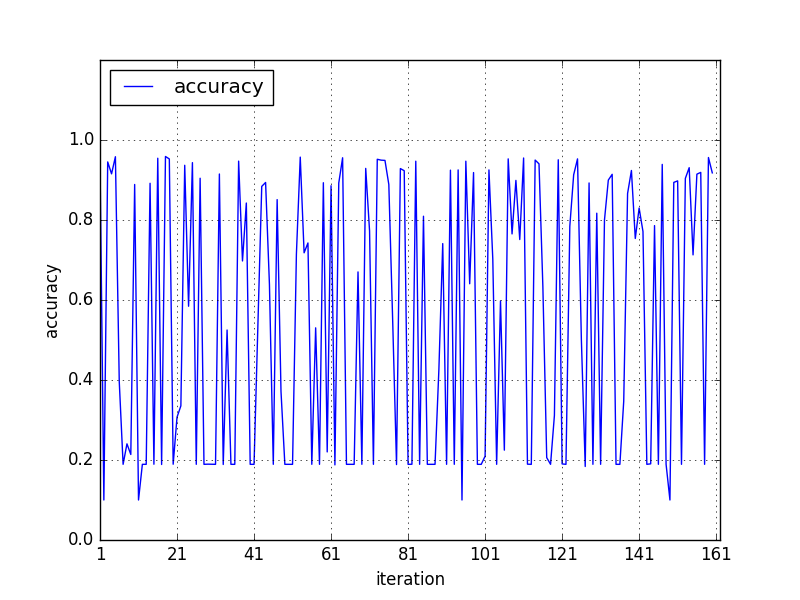}\\
  \end{minipage}}
  \subfigure[]{
  \begin{minipage}{3.8cm}
  \centering
  \includegraphics[height=1.5in,width=1.7in,angle=0]{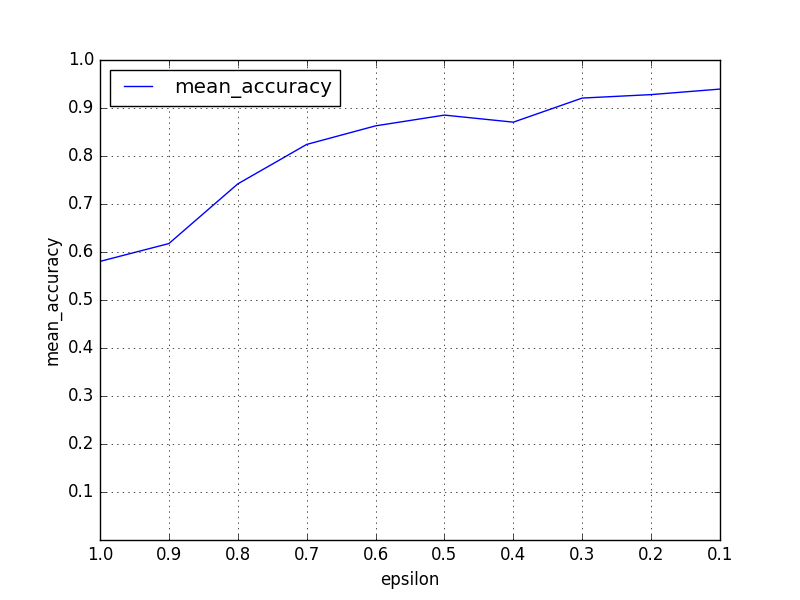}\\
  \end{minipage} }
  \quad
  \subfigure[]{
  \begin{minipage}{4cm}
  \centering
  \includegraphics[height=1.5in,width=1.7in,angle=0]{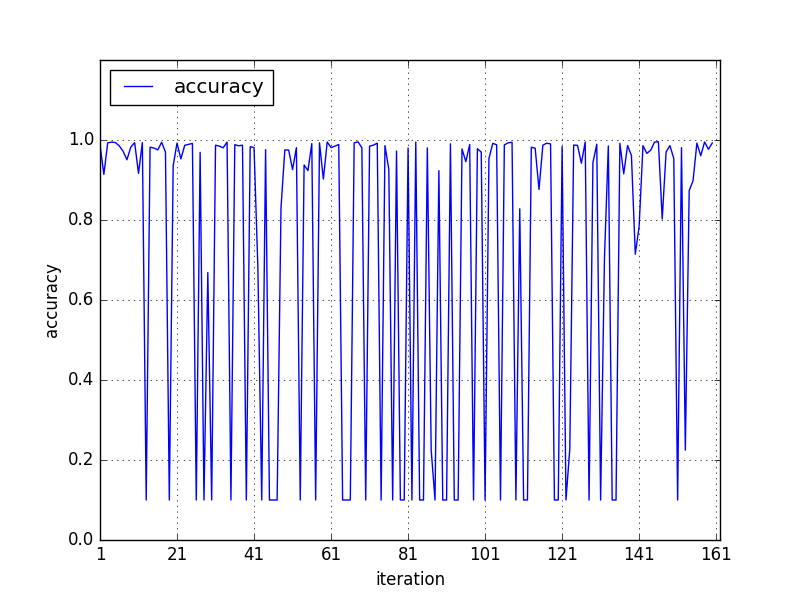}\\
  \end{minipage}}
  \subfigure[]{
  \begin{minipage}{4cm}
  \centering
  \includegraphics[height=1.5in,width=1.7in,angle=0]{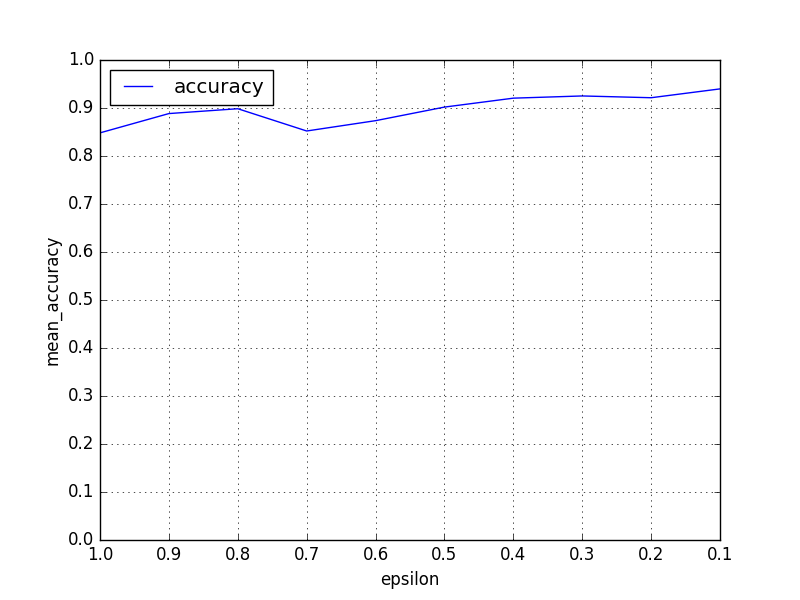}\\
  \end{minipage}}
  \caption{(a), (c), (e) shows the selected model accuracy versus iteration on CIFAR-10, SVHN and MNIST respectively. (b), (d), (f) shows the mean accuracy versus  epsilon stage on CIFAR-10, SVHN and MNIST. }\label{figure5}
\end{figure}

The left column of Figure \ref{figure5} shows the selected model accuracy versus iteration. They might look  a little bit erratic due to the  uncertainty of the combinations of different block modules, so we plotted the mean accuracy during each epsilon stage (the right column of Figure \ref{figure5}). Overall,  the mean accuracy  roughly increases with the epsilon decreases, and the top networks are mostly generated in the final exploitation phrase which can be clearly displayed in Figure \ref{figure5} and Table \ref{table4}, \ref{table5}, \ref{table6}. That is to say, the Q-learning agent gradually improves its ability to select better-structured models  from the exploration period (\(\epsilon\) = 1) to the exploitation stage (0.9 $\le $ \(\epsilon\) $\le $ 0.1 ). For example, the mean accuracy of  models on SVHN dataset increases from 58.04\% at \(\epsilon\) = 1 to 93.94\% at \(\epsilon\) = 0.1.\\
\subsection{Observation and Analysis}
The experimental results suggest that, there are many combinations that different types of block modules can be assembled to form a well-behaved neural network structure. Owing to the difference of  intrinsic structures of these three datasets, the multi-block networks selected by the agent on different datasets are mostly different. Even though some of the multi-block networks picked on three datasets are duplicated, the total number is enough to explain that our method can be effective and efficient. For example, as shown in Table \ref {table4}, the net [B(0), B(6), B(7), SM(10)] is composed of  the starting point dense block B(0) and two different inception blocks B(6), B(7), and the net  [B(0), B(8), B(3), B(0), SM(10)] are made up of two dense blocks B(0), a inception block B(8) and a variant of  residual block B(3).

There are some interesting findings and comparison. (Except item (1), all the networks illustrated in the following were only trained for 30 epochs during the Q-learning process and didn't do any further training.) \\
(1)  One model that performs well in CIFAR-10 dataset might also be selected in another two datasets, and has good performance as well, such as the net [B(0), B(0), SM(10)] selected by the agent with test error rate 4.68\%, 2.44\%, 0.40\% respectively on CIFAR-10, SVHN and MNIST. \\
(2) From our CIFAR-10, SVHN and MNIST replay-database which stores the information of all the models selected by the agent, we find that, as long as the block module B(1) who keeps the original residual form (shown in Figure \ref{figure4}(b)) is one of the component of the constructed network structure, the accuracy is always 0.1 on three datasets, no matter how to combine it  with other block modules. This is also reflected in the curve shown in Figure\ref{figure5} (a), (c), (d), which explains why it appears to have large fluctuations and seems  a little bit irregular. Except B(1),  the rest of other block modules are possible to perform well in a network. From this point we might reach a conclusion that if we want to combine the original residual block from \cite{resnet}  with a dense block to construct a network, the result may be counter-productive. \\
(3)  Swapping the position of the two well-performing block modules that make up a network doesn't lead to much difference in performance. For example, the accuracy of the net [B(0), B(4), B(4), B(3), SM(10)] and the net  [B(0), B(3), B(4), B(4), SM(10)] on MNIST are 99.26\% and 99.20\% respectively. The same instances can also be seen in the replay-database on CIFAR-10 and SVHN dataset. Take the net [B(0), B(6), B(2), SM(10)] with 84.84\% and the net [B(0), B(2), B(6), SM(10)] with 86.54\% on CIFAR-10 as an example.\\
(4) The effect of the concatenation added to the original residual and inception-like form is positive.
\begin{itemize}
  \item With regard to inception-like block modules (B(5), B(6), B(7), B(8), B(9), B(10), B(11)), whether a concatenation is added or not, they all can be an effective element in a well-behaved network. But the inception-like blocks with concatenation in a model can have a little better performance. Examples are selected from replay database for illustration. On SVHN,  the accuracy of the nets  [B(0), B(4), B(11), B(7), SM(10)], [B(0), B(4), B(11), B(8), SM(10)], [B(0), B(4), B(11), B(11), SM(10)] are 18.88\%, 76.52\%, 78.44\% respectively. First three block modules of these nets are the same, so the fourth block B(7), B(8), B(11) makes the nets different in precision. Except that B(7) keeps the original inception form, the B(8), B(11) are the variants that adds concatenation.
  \item As for the residual blocks set by us, from item(2) we have known that stacking the original residual blocks (B(1)) leads to the accuracy of the whole network always 0.1. And from our replay-database, we can see that the concatenation added to the residual units (B(2), B(3), B(4)) really works.
  \item The residual blocks with concatenation perform better than inception-like blocks with concatenation. Take the net [B(0), B(4), B(4), B(3), SM(10)] with 99.26\% and [B(0), B(4), B(4), B(11), SM(10)] with 98.8\% on MNIST as an example. (B(3), B(11) are residual and inception-like blocks with concatenation respectively). The net [B(0), B(4), B(11), B(2), SM(10)] with 92.38\% and the net [B(0), B(4), B(11), B(11), SM(10)] with 78.44\% on SVHN also demonstrate this point.\\
\end{itemize}

\section{Conclusion}
In this paper, we take advantages of the superiority of state-of-the-art networks to automatically generate multi-block neural networks using reinforcement learning method and validate the possibility that different types of blocks can be composed together to form a well-behaved neural network. Our multi-block neural networks can achieve competitive performance with hand-crafted networks as well as other auto-generated networks in image classification problems.

In  the future, it would be possible to search in a larger state-action space to find more outstanding networks with distributed framework and other methods to accelerate convergence. The automatic generation of multiple types of blocks and then the entire neural network will also be a direction of our future research.

{\small
\bibliographystyle{ieee}
\bibliography{mybib}
}

\end{document}